\documentclass[11pt]{article}

\usepackage[]{ACL2023}

\usepackage{times}
\usepackage{latexsym}
\usepackage{covington}
\usepackage[T1]{fontenc}

\usepackage[utf8]{inputenc}

\usepackage{inconsolata}
\usepackage{graphicx}

\usepackage{multirow}
\usepackage[ruled, vlined, linesnumbered]{algorithm2e}
\usepackage{amsmath}
\usepackage{xcolor}
\usepackage{dsttr}
\usepackage{pstricks}
\usepackage{rtrees}

\def\arrow#1{\pspicture[shift=2pt](#1,0)\psline{->}(#1,0)\endpspicture}

\usepackage[normalem]{ulem}
\usepackage{MnSymbol}

\def\paperDraft{}

\ifdefined\paperDraft
 \def\aacomment#1{{\color{aacom}[AA: \textit{#1}]}}
 \def\aecomment#1{{\color{magenta}[AE: \textit{#1}]}}
  \def\aedel#1{{\color{magenta} \sout{#1}}}
   \def\aadel#1{{\color{violet} \sout{#1}}}

\else
 \def\aacomment#1{}
 
 \def\aecomment#1{}
 
  \def\aedel#1{}
  \def\aadel#1{}
\fi
\DeclareMathOperator*{\argmax}{arg\,max}
\newcommand{\meet}{\bigwedgedot}

\definecolor{refcolor}{HTML}{0071BC}
\definecolor{aacom}{HTML}{00B5BE}

\SetKwIF{If}{ElseIf}{Else}{if}{:}{else if}{else}{end if}%
\SetKwFor{While}{while}{:}{end while}%
\SetKwFor{For}{for}{:}{}%

\title{Learning to generate and \texttt{corr-} uh I mean \textit{repair} language in real-time}

\author{Arash Eshghi*$\dagger$~~Arash Ashrafzadeh*\\
  *Heriot-Watt University, Edinburgh, United Kingdom\\
  $\dagger$AlanaAI, Edinburgh, United Kingdom\\
  \texttt{\{a.eshghi, aa2070\}@hw.ac.uk}
}

\begin{document}
\maketitle
\begin{abstract}
In conversation, speakers produce language \textit{incrementally}, word by word, while continuously monitoring the appropriateness of their own contribution in the dynamically unfolding context of the conversation; and this often leads them to repair their own utterance on the fly. This real-time language processing capacity is furthermore crucial to the development of fluent and natural conversational AI. In this paper, we use a previously learned Dynamic Syntax grammar and the CHILDES corpus to develop, train and evaluate a probabilistic model for incremental generation where input to the model is a purely \textit{semantic generation goal concept} in Type Theory with Records (TTR)\footnote{All relevant code, models, and data are available at \url{https://bitbucket.org/dylandialoguesystem/dsttr/src/dsttr_arash_a/}}. We show that the model's output exactly matches the gold candidate in 78\% of cases with a ROUGE-l score of 0.86. We further do a zero-shot evaluation of the ability of the same model to generate \textit{self-repairs} when the generation goal changes mid-utterance. Automatic evaluation shows that the model can generate self-repairs correctly in 85\% of cases. A small human evaluation confirms the naturalness and grammaticality of the generated self-repairs. Overall, these results further highlight the generalisation power of grammar-based models and lay the foundations for more controllable, and naturally interactive conversational AI systems.
\end{abstract}


\section{Introduction}



People process language incrementally, in real-time (see \newcite{Crocker.etal00,Ferreira96,Kempson.etal16} among many others), i.e. both language understanding and generation proceed on a word by word rather than a sentence by sentence, or utterance by utterance basis. This real-time processing capacity underpins participant coordination in conversation \cite{Gregoromichelaki.etal12,Gregoromichelaki.etal20} and leads to many characteristic phenomena such as split-utterances \cite{Poesio.Rieser10,Purver.etal09}, mid-utterance feedback in the form of backchannels \cite{Heldner.etal13} or clarification requests \cite{Healey.etal11,Howes.Eshghi21}, hesitations, self-repairs \cite{Schegloff.etal77} and more.

Language generation -- our focus here -- is just as incremental as language understanding: speakers normally do not have a fully formed conceptualisation or plan of what they want to say before they start articulating, and conceptualisation needs only to be one step ahead of generation or articulation \cite{Guhe07,Levelt89}. This is possible because speakers are able to continuously monitor the syntax, semantics, and the pragmatic appropriateness of their own contribution \cite{Levelt89} in the fast, dynamically evolving context of the conversation. In turn this allows them to pivot or correct themselves on the fly if needed, e.g. because they misarticulate a word, get feedback from their interlocutors \cite{Goodwin81}, or else the generation goal changes due to the dynamics of the environment.

Real-time language processing is likewise crucial in designing dialogue systems that are more responsive, more naturally interactive \cite{Skantze.Hjalmarsson10,Aist.etal06}, and are more accessible to people with memory impairments \cite{addlesee.19, Addlesee.Damonte23, Nasreen.etal21}. Despite this importance, relative to turn-based systems, it has received little attention from the wider NLP community; perhaps because it has deep implications for the architecture of such systems \cite{Schlangen.Skantze09, Skantze.Schlangen09,Kennington.etal14}, which make them much harder to build and maintain.





In this paper, we extend the work of \newcite{Purver.Kempson04,Hough.Purver12,Hough15}, who lay the theoretical foundations for incremental generation and later the processing of self-repairs in Dynamic Syntax \citep[][Sec.~\ref{sec:repair-model}]{Kempson.etal01,Kempson.etal16}. For the first time, we develop a probabilistic model for incremental generation (Sec.~\ref{sec:model}) that conditions next word selection on the current incrementally unfolding context of the conversation, and also on features of a \textit{purely semantic generation goal concept}, expressed as a Record Type (RT) in Type Theory with Records \cite{Cooper12, Cooper.Ginzburg15}. 
The model is trained and evaluated on part of the CHILDES corpus \cite{MacWhinney00} using an extant grammar that was learned by \newcite{Eshghi.etal13a} from the same data. Results show that in the best case, the model output matches the gold generation test candidate in 83\% of cases (Sec.~\ref{sec:vanilla-eval}). We then go on to experiment with and evaluate the ability of the same model to generate self-repairs in a zero-shot setting in the face of \textit{revisions to the goal concept RT} under various conditions (Sec~\ref{sec:repair-eval}): viz. for forward-looking and backward-looking repair and at different distances from the reparandum. 
Automatic evaluation shows that it can generate self-repairs correctly in 85\% of cases. A small human evaluation confirms the overall naturalness and grammaticality of the generated repairs. Overall, these results further highlight the generalisation power of grammar-based models (see also \newcite{Mao.etal21,Eshghi.etal17} and lay the foundations for more controllable, and naturally interactive conversational AI systems.

\section{Dynamic Syntax and Type Theory with Records (DS-TTR)}
\label{sec:DS}
Dynamic Syntax \citep[DS,][]{Kempson.etal16,Cann.etal05a,Kempson.etal01} is a process-oriented grammar formalism that captures the real-time, incremental nature of the dual processes of linguistic comprehension and production, on a word by word or token by token basis. It models the time-linear construction of \emph{semantic} representations (i.e. \emph{interpretations}) as progressively more linguistic input is parsed or generated. DS is idiosyncratic in that it does not recognise an independent level of structure over words: on this view syntax is sets of constraints on the incremental processing of semantic information.

\begin{figure*}[ht]\centering
\vspace{-0.2cm}
\includegraphics[width=\textwidth]{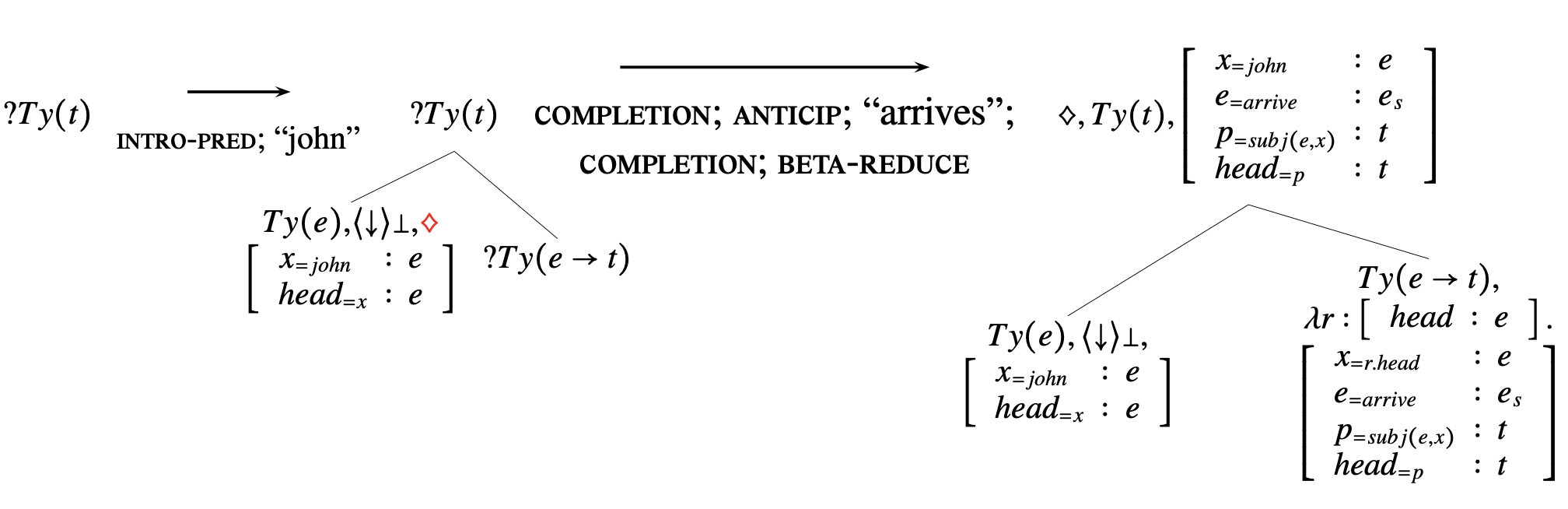}
\caption{Incremental parsing in DS-TTR: \emph{``John arrives''}}\label{fig:tree}\vspace{-0.5cm}
\end{figure*}

The output of parsing any given string of words is thus a \textit{semantic tree} representing its predicate-argument structure (see Fig.~\ref{fig:tree}). DS trees are always binary branching, with argument nodes conventionally on the right and functor nodes to the left; tree nodes correspond to terms in the lambda calculus, decorated with labels expressing
their semantic type (e.g. $Ty(e)$) and formulae -- here as record types of Type Theory with Records (TTR, see Sec.~\ref{sec:ttr} below); and beta-reduction determines the type and formula at a mother node from those at its daughters (Fig.~\ref{fig:tree}). These trees can be \emph{partial}, containing unsatisfied \textit{requirements} potentially for any element (e.g. $?Ty(e)$, a requirement for future development to $Ty(e)$), and contain a \emph{pointer}, $\ptr$, labelling the node currently under development. 

Grammaticality is defined as parsability in a context: the successful incremental word-by-word construction of a tree with no outstanding requirements (a \emph{complete} tree) using all information given by the words in a string. We can also distinguish \emph{potential grammaticality}  (a successful sequence of steps up to a given point, although the tree is not  complete and may have  outstanding requirements) from  \emph{ungrammaticality} (no possible sequence of steps up to a given point).


Fig.~\ref{fig:tree} shows ``John arrives'', parsed incrementally, starting with the axiom tree with one node ($?Ty(t)$), and ending with a complete tree. The intermediate steps show the effects of: (i) DS Computational Actions (e.g. \textsc{Completion} which moves the pointer up and out of a complete node or \textsc{Anticipation} which moves the pointer down from the root to its functor daughter.) which are language-general and apply without any lexical input whenever their preconditions are met; and (ii) Lexical Actions which correspond to words and are triggered when a word is parsed.

\paragraph{Context} In DS, context, required for processing various forms of context-dependency -- including pronouns, VP-ellipsis, and short answers, as well as self-repair -- is the parse search Directed Acyclic Graph (DAG), and as such, is also process-oriented. Edges correspond to DS actions -- both Computational and Lexical Actions -- and nodes correspond to semantic trees after the application of each action \citep{Sato11,Eshghi.etal12,Kempson.etal15}. Here, we take a coarser-grained view of the DAG with edges corresponding to words (sequences of computational actions followed by a single lexical action) rather than single actions, and we drop abandoned parse paths \citep[see][for details]{Eshghi.etal15,Howes.Eshghi21} -- Fig. \ref{fig:dsttrRepair} shows an example.

\subsection{Type Theory with Records (TTR)}
\label{sec:ttr}
Dynamic Syntax is currently integrated with TTR \citep{Cooper12,Cooper05} as the semantic formalism in which meaning representations are couched \citep{Eshghi.etal12,Purver.etal11,Purver.etal10}\footnote{DS models the structural growth of representations and is agnostic to the formalism for semantic representation. As such, it has also been combined with RDF \cite{Addlesee.Eshghi21} and with vector-space representations \cite{Purver.etal21}}.

TTR is an extension of standard type theory, and has been shown to be useful in contextual and semantic modelling in dialogue  \citep[see e.g.][among many others]{Ginzburg12,Fernandez06,Purver.etal10}, as well as the integration of perceptual and linguistic semantics \citep{Larsson13,Dobnik.etal12,Yu.etal17b}. With its rich notions of underspecification and subtyping, TTR has proved crucial for DS research in the incremental specification of content \citep{Purver.etal11,Hough15}; specification of a richer notion of dialogue context \citep{Purver.etal10}; models of DS grammar learning \citep{Eshghi.etal13a}; and models for learning dialogue systems from data \citep{Eshghi.etal17,Kalatzis.etal16,Eshghi.Lemon14}.

\begin{figure}[!ht]
\includegraphics[width=\linewidth]{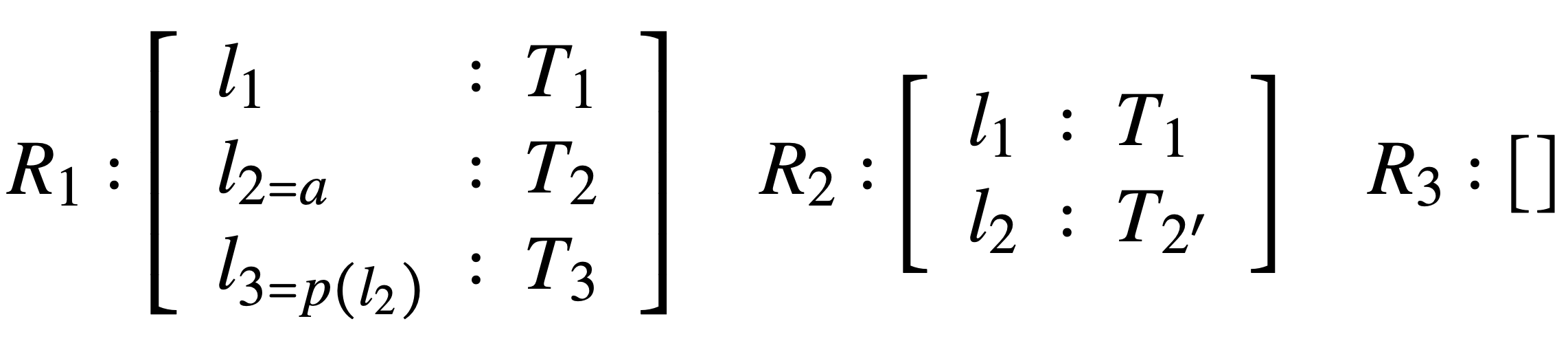}\vspace{-0.1cm}
\caption{Example TTR record types}
\label{fig:ttr}\vspace{-0.1cm}
\end{figure}

In TTR, logical forms are specified as \emph{record types}, which are  sequences of \emph{fields} of the form $\smttrnode{}{l&T}$ containing a label $l$ and a type $T$. Record types can be witnessed (i.e.\ judged true) by \emph{records} of that
type, where a record is a sequence of label-value pairs $\smttrrec{}{l&v}$. We say that $\smttrrec{}{l&v}$ is of type $\smttrnode{}{l&T}$ just in case $v$ is of type $T$. Fields can be \emph{manifest}, i.e.~given a singleton type e.g.~$\smttrnode{}{l & T_{a}}$ where $T_a$ is the type of which only $a$ is a member; here, we write this as $\smttrnode{}{l_{=a}&T}$. Fields can also
be \emph{dependent} on fields preceding them (i.e.~higher) in the record type (see Fig.~\ref{fig:ttr}). 

The standard subtype relation $\subtype$ can be defined for record types: $R_1 \subtype
R_2$ if for all fields $\smttrnode{}{l & T_2}$ in $R_2$, $R_1$ contains $\smttrnode{}{l & T_1}$ where $T_1 \subtype T_2$. In Fig.~\ref{fig:ttr}, $R_1 \subtype R_2$ if $T_2 \subtype T_{2'}$, and both $R_1$ and $R_2$ are subtypes of $R_3$. This subtyping relation allows semantic information to be incrementally specified, i.e. record types can be indefinitely extended with more information and/or constraints. 

\begin{figure}[!ht]\centering\vspace{-0.2cm}
\includegraphics[width=0.85\linewidth]{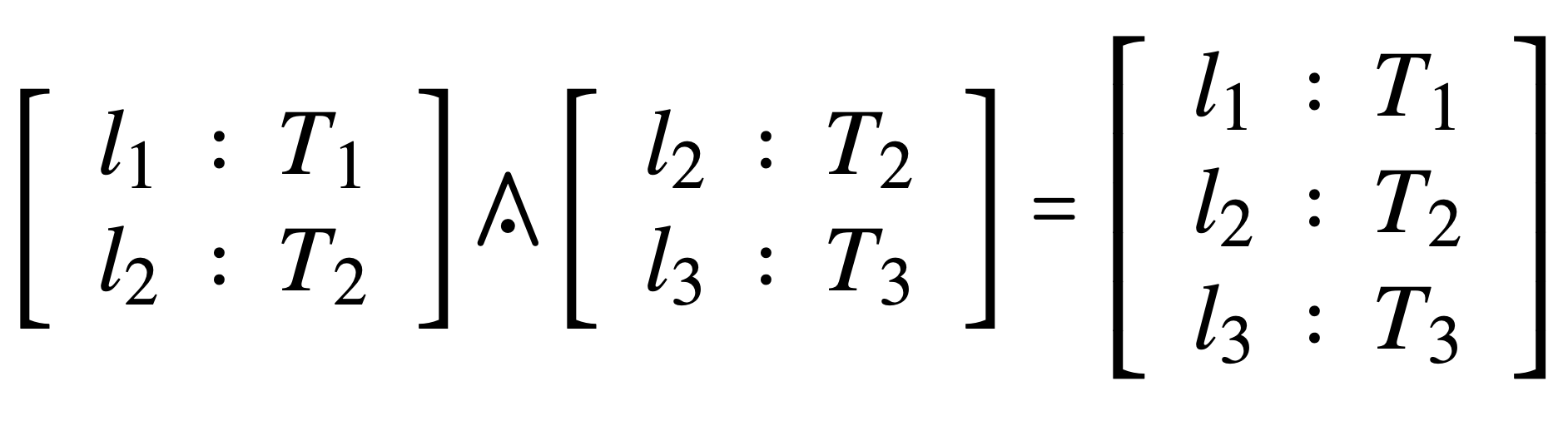}\vspace{-0.3cm}
 \caption{Example of merge operation between two RTs}
 \label{fig:merge}
 \vspace{-.2cm}
 \end{figure}

Additionally, \newcite{Larsson10} defines the meet ($\meet$) operation of two (or more) RTs as the union of their fields; the equivalent of conjunction in FoL; see figure \ref{fig:merge} for an example. We will need this below (Sec.\ref{sec:model}) where we define our probabilistic model.

\vspace{-0.1cm}
\subsection{Generation in DS-TTR}
\label{sec:ds-nlg}
As alluded to in the introduction, to handle typical incremental phenomena in dialogue such as split utterances, interruptive clarification requests or self-repair, any generation model must be as incremental as interpretation: full syntactic and semantic information should be available after generating every word with continual access to the incrementally unfolding context of the conversation \cite{Hough.Purver12,Eshghi.etal15}. In generation, there is an extra requirement on models, namely \textit{representational interchangability} \cite{Eshghi.etal11}: parsing and generation should employ the same mechanisms and use the same kind of representation so that parsing can pick up where generation left off, and vice versa. 


DS-TTR can meet these requirements, because generation employs exactly the same mechanisms as in parsing \cite{Purver.Kempson04} with
the simple addition of a \emph{subsumption check} against a \emph{generation goal concept}, expressed as a Record Type (RT) in TTR (see Sec.~\ref{sec:ttr}); and where this goal concept can be partial (does not need to correspond to a complete sentence), and need only to be one step ahead of the generated utterance so far. This ease of
matching incrementality in both generation and parsing is not matched by other models aiming to reflect incrementality in the dialogue model while adopting relatively conservative grammar
frameworks, some matching syntactic requirements but without incremental semantics \cite{Skantze.Hjalmarsson10}, others matching incremental growth of semantic input but leaving the
incrementality of structural growth unaddressed \cite{Guhe07}.

As such, generation involves \textit{lexical search} whereby at every step, words from the lexicon are test-parsed in order to find words that (i) are parsable in the current context; and (ii) the resulting TTR semantics of the current DS tree subsumes or is monotonically extendable the generation goal. The subsumption relation is the inverse of the subtype relation defined above (see Sec.~\ref{sec:ttr}; i.e. $R_1 subsumes\ R_2$ iff $R_2 \subtype R_1$). 

Without a probabilistic model for word selection at each step of generation, this process is effectively brute-force, computationally very inefficient, and therefore simply impractical, especially with large lexicons. This is the shortcoming that we address here for the first time by conditioning word selection on the generation goal RT. This involves learning, through Maximum Likelihood Estimation from data, $P(w|T, R_g)$, where $w$ ranges over the lexicon, $T$ is the current DS tree including its maximal semantics, and $R_g$ is the generation goal. This parametrisation is described in full below in Sec.~\ref{sec:model}.


\subsection{Processing Self-repair in DS-TTR}
\label{sec:repair}
In this section, we briefly introduce the DS model of self-repair from \cite{Hough.Purver12}: there are two types of self-repair that are addressed: \textit{backward-looking repair} (aka. overt repair), where the repair involves a local, and partial restart of the reparandum, as in (\ref{eg:simplerepair2}) and forward-looking repair (aka. covert repair) where the repair is simply a local extension, i.e. a further specification of the reparandum as in (\ref{eg:simplerepair3}).

\begin{examples}
\item\label{eg:simplerepair2}
``Sure enough ten minutes later the bell r-the doorbell rang"
\emph{\cite{Schegloff.etal77}}
\item\label{eg:simplerepair3}
``I-I mean the-he-they, y'know the guy, the the pathologist, looks at the tissue in the microscope\ldots''
\emph{\cite{Schegloff.etal77}}

\end{examples}

In the model set out above, a backward-looking repair arises due to an online revision of a generation goal RT, whereby the new goal is not a sub-type of the one the speaker (or the dialogue manager) had initially set out to realise. We model this via backtracking along the incrementally available context DAG as set out above. More specifically, repair is invoked if there is no possible DAG extension after the test-parsing and subsumption check stage of generation (resulting in no candidate succeeding word edge). 

\begin{figure*}[!ht]\centering
   \includegraphics[scale=0.35]{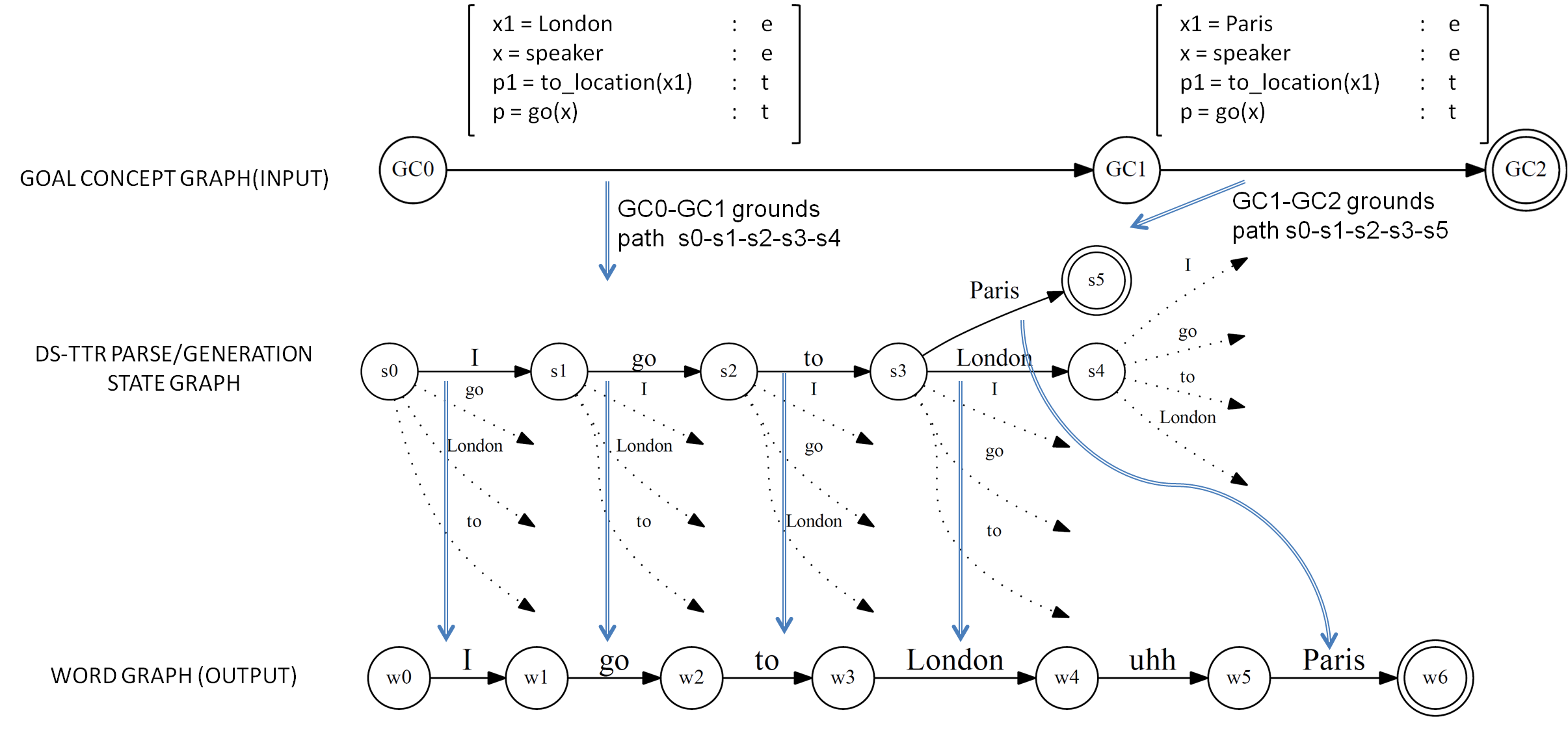} 
   \caption{Incremental DS-TTR generation of a self-repair upon change of goal concept. Type-matched record types are double-circled nodes and edges indicating failed paths are dotted.}
   \label{fig:dsttrRepair}
\end{figure*}

The repair procedure proceeds by restarting generation from the last realised (generated) word edge. It continues backtracking by one DAG vertex at a time until the root record type of the current partial tree is a subtype of the new goal concept. Generation then proceeds as usual by extending the DAG from that vertex. The word edges backtracked over are not removed, but are simply marked as repaired (see also \newcite{Eshghi.etal15} for a fuller account), following the principle that the revision process is on the public conversational record and hence should still be accessible for later anaphoric reference (see Fig.~\ref{fig:dsttrRepair}).

\label{sec:repair-model}


Forward-looking repairs on the other hand, i.e. \emph{extensions}, where the repair effects an ``after-thought'' 
are also dealt with straightforwardly by the model. The DS-TTR parser simply treats these as monotonic extensions of the current tree, resulting in subtype extension of the root TTR record type. Thus, a change in goal concept during generation will not always put demands on the system to backtrack, such as in generating the
fragment after the pause in ``I go to Paris \ldots from London". Backtracking only operates at a semantics-syntax
mismatch where the revised goal concept is no longer a subtype of the root record type for the (sub-)utterance so far realised, as in Figure \ref{fig:dsttrRepair}.


\section{Probabilistic Model of Generation}
\label{sec:model}
In this section, we follow on from Sec.~\ref{sec:repair-model} above and describe the probabilistic model that we have developed for incremental probabilistic generation. First we describe the model itself, its parameters, and how these are estimated from data. Then we describe how the model is used at inference time to generate.

\paragraph{Model and Parameter Estimation} As noted, generation in Dynamic Syntax is defined in terms of parsing. Specifically, it proceeds via lexical search, i.e. test-parsing (all) words from the lexicon while checking for \textit{subsumption} against the \textit{goal concept}: a record type (RT) in TTR; henceforth $R_g$. Words that parse successfully with a resulting (partial) semantics that subsume the goal concept are successfully generated. This process goes on until the semantics of the generated sentence equals the goal. This process is highly inefficient and impractical for larger lexicons.

On a high level, we solve this problem by building a probabilistic model which conditions the probability of generating the next word, $w$, on: (i) $R_{cur}$: the semantics of the generated utterance thus far; (ii) $R_g$, the goal concept; and (iii) the current DS tree (henceforth $T_{cur}$). We condition on (i) to allow the model to keep track of the semantics of what's already been generated, i.e. the left semantic context of generation; on (ii) to aid the model in selecting words that contribute the correct semantic increments to approach the goal concept; and on (iii) to capture the syntactic constraints on what words can grammatically follow. In sum, we need to compute $P(w|T_{cur}, R_{cur}, R_{g})$ for all the words $w$ in the lexicon.

As you will see below, we learn to generate by parsing, and therefore we the use Bayes rule in Eq.~\ref{eq:main} to cast probabilistic generation roughly in probabilistic parsing terms:
\begin{equation}
    \begingroup\color{refcolor}\underbrace{\color{black}P(w|T_{cur}, R_{cur}, R_{g})}_\text{\color{black}probabilistic generation}\endgroup \stackrel{\text{Bayes Rule}}{=} \dfrac{\begingroup\color{refcolor}\overbrace{\color{black}P(T_{cur}, R_{cur}, R_{g}|w)}^\text{\color{black}probabilistic parsing}\endgroup P(w)}{P(T_{cur}, R_{cur}, R_g)}
\end{equation}\label{eq:main}

On the right hand side of Eq.~\ref{eq:main}, $P(w)$ is the prior probability of $w$, which we obtain from the frequency of $w$ in our training data; and $P(T_{cur}, R_{cur}, R_g)$ a normalisation constant which we do not need to estimate.

We learn $P(T_{cur}, R_{cur}, R_{g}|w)$ from gold data in the form of $\langle Utt = \langle w_1,\dots,w_N\rangle,\ R_g\rangle$, where $Utt$ is the utterance to be generated, and $R_g$ is its gold semantics. To do this, we use the DS parser to parse $Utt$ yielding a parse path (see e.g. Fig.~\ref{fig:dsttrRepair}) that starts with the DS axiom tree (empty tree) to the tree whose semantics is $R_g$ together with all the DS trees produced after parsing each $w_i$ in between; viz. a sequence $S_p = \{\langle T_1, w_1\rangle,\dots,\langle T_N, w_N\rangle\}$, where $T_i$ are the DS trees in the context of which the $w_i$'s were parsed. This sequence constitutes the observations from which we estimate $P(T_{cur}, R_{cur},R_g|w)$ by Maximum Likelihood Estimation (MLE).

$T_{cur}$, $R_{cur}$ and $R_g$ are all composed of many individual features, and as a whole, would be observed very rarely. Therefore, for generalisation, we need to decompose them and compute the probability of the whole as the conjunction (product) of the probabilities of their individual atomic features. 

For $T_{cur}$ we follow \newcite{Eshghi.etal13a} and consider only one feature of $T_{cur}$: that of the type of the pointed node, or a requirement for a type (e.g. $Ty(e)$, $?Ty(e\rightarrow t)$, etc) -- call this $Ty_p$. This simplifies the model considerably, but has the downside of not capturing all grammatical constraints (e.g. \textit{person} constraints in English verbs will not be captured this way), and leading to some over-generation.

We also simplify the model by conditioning on the semantics that \textit{remains to be generated} -- call it $R_{inc}$ -- rather than conditioning on both $R_{cur}$ and $R_g$. We can compute $R_{inc}$ each time through the well-defined \textit{record type subtraction} operation in TTR where: $R_{inc} = R_g \backslash R_{cur}$.

With these simplifications, what we need to estimate by MLE from each sequences $S_p$ (see above) is: $P(Ty_p, R_{inc}|w)$.

As noted, for any generalisation at all, $R_{inc}$ now needs to be decomposed into its individual atomic features so that we can compute the probability of each of these features individually, rather than that of $R_{inc}$ as a whole. We decompose $R_{inc}$ as follows: $R_{inc} = \meet_k (R_k)$, where $\meet$ is the TTR equivalent of the conjunction operation in FoL (see above, Sec.~\ref{sec:ttr}); and each $R_k$ is potentially \textit{dependent} on $R_j$ where $j<k$.

Using the probabilistic variant of TTR \cite{Cooper.etal13ProbTTR,Cooper.etal14ProbTTR}, we can use the chain rule to then derive Eq.~\ref{eq:expansion2}:
\vspace{-0.2cm}
\begin{equation}
    P(\meet_k R_{k}| w) = \Pi_{k} P(R_k| w, R_1\meet \dotsc \meet R_{k-1})
    \label{eq:expansion2}
\end{equation}

This then allows us to express the probability of a complex record type in terms of the product of its potentially \textit{dependent}, atomic supertypes. This, finally, puts us in a position to compute $P(Ty_p, R_{inc}|w)$ as follows:

\begin{align*}
\small
    P(R_{inc}, Ty_p|w) & \stackrel{independence}{=} P(R_{inc}| w)\cdot P(Ty_p|w)\\ &\stackrel{decompose\ R_{inc}}{=} P(\meet_k R_k| w)\cdot P(Ty_p|w)
\end{align*}

We implement the above procedure by constructing a 2D conditional count table where the rows are the words, and the columns are all the atomic semantic features observed during learning by parsing: effectively the result of decomposing all the $R_g$'s in our data; this, in addition to all the $Ty_p$ features we have observed on all the DS trees encountered in the $S_p$ sequences above. Then, each time we observe an atomic semantic feature of $R_{inc}$, say, $R_k$, in the context of a word, $w$, we increment the cell $(R_k,w)$ by 1. After learning, we normalise the columns of the table to obtain all $P(F|w)$ where $F$ ranges over all semantic features and pointed node types, and $w$ over all words in the lexicon.

\paragraph{Inference} At inference time, we need to estimate $P(w|T_{cur}, R_{cur}, R_{g})$: a probability distribution over all the words in the lexicon, given the current context of generation, $T_{cur}$ including the current semantics so far generated, $R_{cur}$, and the goal concept, $R_g$. Given the above we take the following steps to \textit{populate a beam} for generating the next word: (i) compute $R_{inc} = R_g \backslash R_{cur}$; (ii) compute all the atomic semantic features, $R_k$ -- the headings in the columns in our conditional probability table -- that $R_{inc}$ triggers or `turns on'. This can be done by checking whether $R_{inc} \subtype R_k$; (iii) compute the single $Ty_p$ (type of pointed node) feature by observing the type of the pointed node on $T_{cur}$; (iv) for each row (i.e. each word) take the product (or sum of log probabilities) of all the column features thus triggered in steps (ii) and (iii); (v) sort the words in the lexicon by their probability from (iv) and have the top N fill the beam of size N.

Once the beam is thus populated, we use the DS grammar to parse each word in the beam in turn; upon success, that is, if the word is parsable, and the resulting semantics subsumes the goal concept, $R_g$, we move on to generate the next word incrementally until we reach the goal concept, that is, until $R_g \subtype R_{cur} \wedge R_{cur} \subtype R_g$.

\paragraph{Repair mechanism} The DS repair mechanism, i.e. that of backtrack and parse / generate (see above Sec.~\ref{sec:repair-model}), is triggered when none of the words in the beam successfully generate; either because neither are parsable, or else their resulting semantics don't subsume $R_g$ (because it may have been revised). When triggered, the model backtracks over the context DAG path (see above), and, following the same inference process, attempts to (re-)populate the beam and generate from there. Backtracking continues until generation is successful, with the model having generated the interregnum (e.g. "I mean", "sorry I mean", "uh", "no", etc.) before it generates the first repair word. Generation continues normally from that point until the (potentially new) goal concept is reached.

\section{Evaluation}
\subsection{Data}
The data to train and test our model comes from the Eve section of the CHILDES corpus \citep{MacWhinney00}. This section was annotated with logical forms (LF) by \newcite{Kwiatkowski.etal12}. The LFs were then converted to TTR record types (RT) by \newcite{Eshghi.etal13a}. This dataset consists of utterances towards children from parents, therefore the sentences have a relatively simple structure than adult language. We will use it in the shape of $\langle$Utterance, Goal Concept$\rangle$ pairs to train and test our model.

For training our generator, we test-parsed the dataset using two versions of the grammar learned by \newcite{Eshghi.etal13a}: the grammar containing the top 1 hypothesis and the grammar containing the top 3. This resulted in two subsets of the data that could be parsed and in which the produced RT semantics matched the gold semantics exactly. Let's call these \texttt{top-1} and \texttt{top-3} respectively. We report their relevant statistics in Table~\ref{tab:stats}.

\begin{table}[ht]\centering\footnotesize
\begin{tabular}{|c|p{1.1cm}|p{0.7cm}|p{0.7cm}|p{0.7cm}|p{0.8cm}|}\hline
dataset & total samples & total words & mode length & max length & type / token ratio\\\hline
top-1 & 729 & 2152 & 3 & 7 & 18.08\\\hline
top-3 & 1361 & 4194 & 3 & 7 & 21.96\\\hline
\end{tabular}\caption{Filtered Dataset Statistics}\label{tab:stats}\vspace{-0.4cm}
\end{table}

However, even as the top-3 grammar from \newcite{Eshghi.etal13a} gives wider parsing coverage, it included many erroneously learned lexical actions. We therefore decided to carry out our experiments below on the \texttt{top-1} dataset filtered using the top-1 grammar. This is at the expense of not generating sentences that we'd otherwise be able to generate since the overall distribution of the two datasets are similar. Therefore, the results we report below are more conservative (i.e. lower) than those we'd have been able to achieve if we'd manually cleaned up the top-3 grammar and applied it to learning and generation.

\label{sec:eval}

\subsection{Model Evaluation}
\label{sec:vanilla-eval}
We evaluate our generation model on the \texttt{top-1} set in two ways: (i) standard evaluation of generation without any mid-generation revisions to the goal; (ii) we evaluate the capability of the same model to generalise to cases where the goal concept is revised mid-generation, i.e. to cases where the model needs to produce \textit{self-repairs}.

 \paragraph{Standard evaluation} For this, we report percentage of exact match (EM), ROUGE-1, Rouge-2, and ROUGE-l 
 between the gold sentences in the dataset and the output sentences from the model. On the training set, we could observe that out of 656 training samples, we can generate 597 utterances (91.01\%) whose semantics exactly matches the generation goal concept; 416 of these fully match the gold sentence, yielding an EM score of 0.6341 (meaning 63.41\% of the output sentences fully match the gold sentences). For the test set, out of 73 total samples, 64 sentences were generated fully to the goal concept (87.67\%), and 46 of these (63.01\%) completely matched the gold sentence in the dataset. Among the outputs not fully match by the gold sentences a large portion of them are very close to an exact match. For example the generated sample, ``what is that", where the gold sentence is ``what's that": such samples were not counted initially among the exact matches. We then took these to be exact matches and recomputed evaluation scores. The final results are summarised in Table.~\ref{tab:vanilla-eval}.

\begin{table}[ht]\footnotesize
\begin{tabular}{|p{0.7cm}|c|c|c|c|c|}
\hline 
   & EM & ROUGE-1 & ROUGE-2 & ROUGE-l \\ \hline 
Train & 0.84 & 0.94 & 0.71 & 0.92 \\\hline 
Test & 0.78 & 0.88 & 0.67 & 0.86 \\\hline 
\end{tabular}

\caption{Evaluation results for generation without any goal concept revisions}
\label{tab:vanilla-eval}
\end{table}




\vspace{-0.8cm}
\subsection{Generating self-repairs: a zero-shot evaluation}
\label{sec:repair-eval}
To evaluate the ability of the model to generate self-repairs in a zero shot setting, we generate a dataset of \textit{semantic revisions} to the goal concept using the original \texttt{top-1} data. We use the Stanford POS tagger to automatically generate a set of revisions, where each revision is a tuple, $\langle R_g, index, R_r, Utt_r, forward\rangle$: \textbf{$R_g$}: is the original goal concept; \textbf{$index$}: is the position along the generation path where the revision takes place; \textbf{$R_r$}: is the revised goal; \textbf{$Utt_r$}: is the result of replacing a single word in the \textit{original} gold utterance with a word from our data of the same POS -- $R_r$ now corresponds to the (revised goal) semantics of $Utt_r$; and, finally: \textbf{$forward$}: is either true or false, marking whether the revised semantic material has already been contributed before $index$ or not; if true, we would expect a \textit{forward-looking} self-repair, and otherwise a \textit{backward-looking} one (see Sec.~\ref{sec:repair-model} above). We derive these revision tuples for every utterance in the dataset with length greater than 4, and on the following Parts of Speech: \{NOUN, ADJ, PROPN, ADP, ADV\}. These tuples therefore give us 4 experimental conditions, across two binary factors: (i) locality: is the point at which the revision is made strictly local to the repairandum; or does it have a distance of more than 1; (ii) Is the revision after or before the corresponding semantic contributions have been made?


We then run the revisions through the model and evaluate the output automatically as follows: we use a simple rule-based algorithm to `clean out' the self-repair from the model output, and compare this to the revised utterance, $Utt_r$. For this comparison, we only report EM -- see Table~\ref{tab:zero-eval}. We observed 641 of the generatable revisions in total are an exact match.

\begin{table}[ht]
\begin{tabular}{|c|c|c|}
\hline
 & forward-looking & backward-looking \\ \hline
local & 0.93 &  0.89 \\\hline
distant & 0.73 & 0.82\\\hline
\end{tabular}%
\caption{EM for zero-shot evaluation of repairs}
\label{tab:zero-eval} \vspace{-0.5cm}
\end{table}

Since we do not have gold data for self-repairs, we did a small human evaluation on the model output: the authors each independently annotated a subset of 30 examples, assigning scores on a Likert scale from 1 to 3 for: (a) grammaticality of the self-repairs; and (b) their human-likness or naturalness, which initially led to a low agreement. They then met to discuss the disagreements in order to iron out the differences between the criteria they had applied. They then continued to annotate 70 additional system outputs. This led to a Krippendorff's alpha score of 0.88 for grammaticality and 0.82 for naturalness, demonstrating very high agreement. To then report the average scores given by the human annotators, the lower score was chosen when there was a disagreement, resulting in 2.72 and 2.28 mean scores for grammaticality and naturalness respectively, confirming the quality of the generated output.

\section{Discussion}
During the error analysis we observed the following error patterns: In the standard evaluation of generation, there were 199 instances where the model had fully generated to the goal concept, while the generated output did not match the gold utterance. Many were cases where the model had generated a statement instead of a question or vice versa (e.g. "I may see them" is generated over "may I see them"). In a few cases, the generated output was ungrammatical with the wrong word order: both of these are caused by the original grammar from \newcite{Eshghi.etal13a} overgenerating -- this is acknowledged by the authors, and it is due to the fact that their induced grammar did not capture the full set of syntactic constraints present in their data. This is in turn because they were only conditioning their search on the type of the pointed node, like we do here. Inducing the full set of syntactic constraints was left to future work, as it is here.

\subsection{Limitations}
Our evaluation in this paper has at least two important \textbf{limitations}:

(1) We evaluate our incremental generation model on a small, and relatively simple dataset (leading to high ROUGE scores because of the little variation in data and relative similarity between training and testing sets) due to the fact that we currently do not have access to a wider coverage grammar.
However, this was a conscious choice on the authors' part: we used a learned grammar to induce our probabilistic generation model and evaluated it on exactly the same dataset from which the grammar was learned \cite{Eshghi.etal13a}. This was deemed to be methodologically both sounder and cleaner than, say, use of a manually constructed grammar. We also believe that the probabilistic model we have contributed here will generalise to larger, more complex datasets when wider-coverage grammars becomes available. We leave this for future work.

 (2) Perhaps more importantly, we have no comparative evaluation, and this in a climate where neural NLG has seen astonishing advances in the work on Transformer-based (large) Language Models. To carry out this comparative evaluation, we need to integrate our model with a downstream, and, ideally, multimodal dialogue task (see e.g. \newcite{Yu.etal16, Yu.etal17b} for how DS-TTR can be integrated within a visually grounded task). This requires substantial further work which is our next step.

\subsection{Why a grammar-based approach?}
It might reasonably be asked why we are using a grammar-based approach in the age of Large Language Models (LLM) such as GPT-4 and a large number of other, open source models following. These models are astonishing few-shot learners, and have recently achieved great successes that few thought possible (e.g. in open-domain dialogue, conversational question answering, essay writing, summerisation, translation etc), and are changing the human world in ways that we have not yet had time to grasp.

Nevertheless, for the moment, the fact remains that: (a) these models are extremely costly to train and run due their sheer size and the amount of resources (data, compute power, energy) needed to train them; it's also been demonstrated, time and again, that they have poor compositional generalisation properties (see \newcite{Pantazopoulos.etal22,Nikolaus.etal19} among others), which explains much of their characteristic data inefficiency; (b) they are very difficult to \textit{control} and/or adapt while often producing factually incorrect statements, commonly referred to as hallucinations \cite{Rashkin.etal21, Dziri.etal22} using very convincing language -- this extends to confident prediction of erroneous actions or plans in multi-modal, embodied settings; (d) they are very hard to sufficiently \textit{verify}, making them unsuitable for use in safety-critical domains such as healthcare; (e) particularly important for us here, unlike recurrent models such as RNNs and LSTMs, standard Transformer-based neural architectures \cite{Vaswani.etal17} are not properly incremental -- even the auto-regressive variants such as GPT -- in the sense that they process word sequences as whole, rather than word by word; they can be run under an `incremental interface' \cite{Madureira.Schlangen20,Rohanian.Hough21} where input is reprocessed from the beginning with every new token, but even then, they exhibit poor incremental performance with unstable output compared to e.g. LSTMs \cite{Madureira.Schlangen20}. Interesting recent work has explored using Linear Transformers \cite{Katharopoulos.etal20} with recurrent memory to properly incrementalise LMs \cite{Kahardipraja.21}, but this work is as yet in its infancy, and we do not yet know of any work that integrates LMs end to end within a real-time, incremental dialogue system.

On the other hand, grammar-based approaches have the advantage of being highly controllable and transparent; but crucially, they incorporate the very large wealth of linguistic knowledge that has arisen from decades of linguistics and semantics research. This knowledge has been demonstrated to be a very effective source of inductive bias in grammar-based models which in turn translates to remarkable generalisation potential, and thus also data efficiency (see e.g. \newcite{Mao.etal21} for a CCG-based multi-modal model, and \newcite{Eshghi.etal17} for a DS-TTR-based one) -- see \newcite{Eshghi.etal22} for an extended discussion. One common criticism is that grammar-based models are brittle. This is often true, but we do not believe this to be a fundamental property, and think that specific grammars of a language are adaptable and learnable from interaction. But much work remains to be done to demonstrate this property.

For these reasons, we believe that grammar-based approaches hold promises that are as yet unfulfilled, and are therefore still worth exploring in parallel to the much needed work on making LM architectures and training regimes more incremental (see \newcite{Kahardipraja.etal21,Kahardipraja.etal23}).

\section{Conclusion}
We developed the first semantic, probabilistic model of real-time language generation using the Dynamic Syntax framework. The results show that the model performs well, even though we evaluated it only on a small dataset. We also demonstrated the zero-shot generalisation ability of the model to generate self-repairs where none were observed during training. To our knowledge, this is the first model capable of reacting to real-time changes to the generation goal by generating suitable self-corrections. This ability is essential in dialogue systems in highly dynamic contexts or environments. Our generation model can be seamlessly integrated into incremental dialogue system architectures (e.g. based on \newcite{Schlangen.Skantze09}). This work further highlights the generalisation power of grammar-based approaches, and lays the foundations for creating conversational AI systems that are controllable, data-efficient, more naturally interactive, and more accessible to people with cognitive impairments.


\section*{Acknowledgements}
We are very grateful to Tanvi Dinkar and Julian Hough for some of the ideas in this paper and subsequent discussion. We would also like to thank the SemDial reviewers whose constructive critique lead to further changes and elaboration.

\bibliography{anthology,custom,all}

\begin{thebibliography}{67}
\expandafter\ifx\csname natexlab\endcsname\relax\def\natexlab#1{#1}\fi

\bibitem[{Addlesee and Damonte(2023)}]{Addlesee.Damonte23}
Angus Addlesee and Marco Damonte. 2023.
\newblock Understanding and answering incomplete questions.
\newblock In \emph{Proceedings of the 5th Conference on Conversational User
  Interfaces}.

\bibitem[{Addlesee and Eshghi(2021)}]{Addlesee.Eshghi21}
Angus Addlesee and Arash Eshghi. 2021.
\newblock \href {https://aclanthology.org/2021.reinact-1.1} {Incremental
  graph-based semantics and reasoning for conversational {AI}}.
\newblock In \emph{Proceedings of the Reasoning and Interaction Conference
  (ReInAct 2021)}, pages 1--7, Gothenburg, Sweden. Association for
  Computational Linguistics.

\bibitem[{Addlesee et~al.(2019)Addlesee, Konstas, and Eshghi}]{addlesee.19}
Angus Addlesee, Ioannis Konstas, and Arash Eshghi. 2019.
\newblock Current challenges in spoken dialogue systems and why they are
  critical for healthcare applications.
\newblock In \emph{Proceedings of the Dialogue for Good (DiGo) 2019 workshop}.

\bibitem[{Aist et~al.(2006)Aist, Allen, Campana, Galescu, Gomez~Gallo, Stoness,
  Swift, and Tanenhaus}]{Aist.etal06}
G.S. Aist, J.~Allen, E.~Campana, L.~Galescu, C.A. Gomez~Gallo, S.~Stoness,
  M.~Swift, and M.~Tanenhaus. 2006.
\newblock Software architectures for incremental understanding of human speech.
\newblock In \emph{Proceedings of the International Conference on Spoken
  Language Processing (ICSLP)}, Pittsburgh.

\bibitem[{Cann et~al.(2005)Cann, Kempson, and Marten}]{Cann.etal05a}
Ronnie Cann, Ruth Kempson, and Lutz Marten. 2005.
\newblock \emph{The Dynamics of Language}.
\newblock Elsevier, Oxford.

\bibitem[{Cooper(2005)}]{Cooper05}
Robin Cooper. 2005.
\newblock Records and record types in semantic theory.
\newblock \emph{Journal of Logic and Computation}, 15(2):99--112.

\bibitem[{Cooper(2012)}]{Cooper12}
Robin Cooper. 2012.
\newblock Type theory and semantics in flux.
\newblock In Ruth Kempson, Nicholas Asher, and Tim Fernando, editors,
  \emph{Handbook of the Philosophy of Science}, volume 14: Philosophy of
  Linguistics, pages 271--323. North Holland.

\bibitem[{Cooper et~al.(2013)Cooper, Dobnik, Lappin, and
  Larsson}]{Cooper.etal13ProbTTR}
Robin Cooper, Simon Dobnik, Shalom Lappin, and Staffan Larsson. 2013.
\newblock Probabilistic {T}ype {T}heory and {N}atural {L}anguage {S}emantics.
\newblock \emph{Unpublished Manuscript, University of Gothenburg and King's
  College London}.

\bibitem[{Cooper et~al.(2014)Cooper, Dobnik, Lappin, and
  Larsson}]{Cooper.etal14ProbTTR}
Robin Cooper, Simon Dobnik, Shalom Lappin, and Staffan Larsson. 2014.
\newblock A probabilistic rich type theory for semantic interpretation.
\newblock In \emph{Proceedings of the {EACL} Workshop on Type Theory and
  Natural Language Semantics ({TTNLS})}, Gothenburg, Sweden. Association for
  Computational Linguistics.

\bibitem[{Cooper and Ginzburg(2015)}]{Cooper.Ginzburg15}
Robin Cooper and Jonathan Ginzburg. 2015.
\newblock Type theory with records for natural language semantics.
\newblock \emph{The Handbook of Contemporary Semantic Theory}, pages 375--407.

\bibitem[{Crocker et~al.(2000)Crocker, Pickering, and Clifton}]{Crocker.etal00}
Matthew Crocker, Martin Pickering, and Charles Clifton, editors. 2000.
\newblock \emph{Architectures and Mechanisms in Sentence Comprehension}.
\newblock Cambridge University Press.

\bibitem[{Dobnik et~al.(2012)Dobnik, Cooper, and Larsson}]{Dobnik.etal12}
Simon Dobnik, Robin Cooper, and Staffan Larsson. 2012.
\newblock Modelling language, action, and perception in type theory with
  records.
\newblock In \emph{Proceedings of the 7th International Workshop on Constraint
  Solving and Language Processing (CSLP?{\"A}{\^o}12)}, pages 51--63.

\bibitem[{Dziri et~al.(2022)Dziri, Milton, Yu, Zaiane, and
  Reddy}]{Dziri.etal22}
Nouha Dziri, Sivan Milton, Mo~Yu, Osmar Zaiane, and Siva Reddy. 2022.
\newblock \href {https://doi.org/10.18653/v1/2022.naacl-main.387} {On the
  origin of hallucinations in conversational models: Is it the datasets or the
  models?}
\newblock In \emph{Proceedings of the 2022 Conference of the North American
  Chapter of the Association for Computational Linguistics: Human Language
  Technologies}, pages 5271--5285, Seattle, United States. Association for
  Computational Linguistics.

\bibitem[{Eshghi et~al.(2015)Eshghi, Howes, Gregoromichelaki, Hough, and
  Purver}]{Eshghi.etal15}
A.~Eshghi, C.~Howes, E.~Gregoromichelaki, J.~Hough, and M.~Purver. 2015.
\newblock Feedback in conversation as incremental semantic update.
\newblock In \emph{Proceedings of the 11th International Conference on
  Computational Semantics (IWCS 2015)}, London, UK. Association for
  Computational Linguisitics.

\bibitem[{Eshghi et~al.(2011)Eshghi, Purver, and Hough}]{Eshghi.etal11}
A.~Eshghi, M.~Purver, and Julian Hough. 2011.
\newblock Dylan: Parser for dynamic syntax.
\newblock Technical report, Queen Mary University of London.

\bibitem[{Eshghi et~al.(2022)Eshghi, Gregoromichelaki, and
  Howes}]{Eshghi.etal22}
Arash Eshghi, Eleni Gregoromichelaki, and Christine Howes. 2022.
\newblock Action coordination and learning in dialogue.
\newblock In Jean-Philippe Bernardy, Rasmus Blanck, Stergios Chatzikyriakidis,
  Shalom Lappin, and Aleksandre Maskharashvili, editors, \emph{Probabilistic
  Approaches to Linguistic Theory}, chapter~10. CSLI.

\bibitem[{Eshghi et~al.(2013)Eshghi, Hough, and Purver}]{Eshghi.etal13a}
Arash Eshghi, Julian Hough, and Matthew Purver. 2013.
\newblock Incremental grammar induction from child-directed dialogue
  utterances.
\newblock In \emph{Proceedings of the 4th Annual Workshop on Cognitive Modeling
  and Computational Linguistics (CMCL)}, pages 94--103, Sofia, Bulgaria.
  Association for Computational Linguistics.

\bibitem[{Eshghi et~al.(2012)Eshghi, Hough, Purver, Kempson, and
  Gregoromichelaki}]{Eshghi.etal12}
Arash Eshghi, Julian Hough, Matthew Purver, Ruth Kempson, and Eleni
  Gregoromichelaki. 2012.
\newblock \href
  {http://www.eecs.qmul.ac.uk/~mpurver/papers/eshghi-et-al12cooper.pdf}
  {Conversational interactions: Capturing dialogue dynamics}.
\newblock In S.~Larsson and L.~Borin, editors, \emph{From Quantification to
  Conversation: Festschrift for Robin Cooper on the occasion of his 65th
  birthday}, volume~19 of \emph{Tributes}, pages 325--349. College
  Publications, London.

\bibitem[{Eshghi and Lemon(2014)}]{Eshghi.Lemon14}
Arash Eshghi and Oliver Lemon. 2014.
\newblock {How domain-general can we be? Learning incremental dialogue systems
  without dialogue acts}.
\newblock In \emph{Proceedings of {S}emdial 2014 (Dial{W}att)}.

\bibitem[{Eshghi et~al.(2017)Eshghi, Shalyminov, and Lemon}]{Eshghi.etal17}
Arash Eshghi, Igor Shalyminov, and Oliver Lemon. 2017.
\newblock \href {https://doi.org/10.18653/v1/D17-1236} {Bootstrapping
  incremental dialogue systems from minimal data: the generalisation power of
  dialogue grammars}.
\newblock In \emph{Proceedings of the 2017 Conference on Empirical Methods in
  Natural Language Processing}, pages 2220--2230, Copenhagen, Denmark.
  Association for Computational Linguistics.

\bibitem[{Fern{\'a}ndez(2006)}]{Fernandez06}
Raquel Fern{\'a}ndez. 2006.
\newblock \emph{Non-Sentential Utterances in Dialogue: Classification,
  Resolution and Use}.
\newblock Ph.D. thesis, King's College London, University of London.

\bibitem[{Ferreira(1996)}]{Ferreira96}
Victor Ferreira. 1996.
\newblock Is it better to give than to donate? {S}yntactic flexibility in
  language production.
\newblock \emph{Journal of Memory and Language}, 35:724--755.

\bibitem[{Ginzburg(2012)}]{Ginzburg12}
Jonathan Ginzburg. 2012.
\newblock \emph{The Interactive Stance: Meaning for Conversation}.
\newblock Oxford University Press.

\bibitem[{Goodwin(1981)}]{Goodwin81}
C.~Goodwin. 1981.
\newblock \emph{Conversational organization: Interaction between speakers and
  hearers}.
\newblock Academic Press, New York.

\bibitem[{Gregoromichelaki et~al.(2012)Gregoromichelaki, Cann, and
  Kempson}]{Gregoromichelaki.etal12}
Eleni Gregoromichelaki, Ronnie Cann, and Ruth Kempson. 2012.
\newblock \href {https://doi.org/doi:10.1515/tlr-2012-0020} {Language as tools
  for interaction: Grammar and the dynamics of ellipsis resolution}.
\newblock \emph{The Linguistic Review}, 29(4):563--584.

\bibitem[{Gregoromichelaki et~al.(2020)Gregoromichelaki, Mills, Howes, Eshghi,
  Chatzikyriakidis, Purver, Kempson, Cann, and
  Healey}]{Gregoromichelaki.etal20}
Eleni Gregoromichelaki, Gregory~James Mills, Christine Howes, Arash Eshghi,
  Stergios Chatzikyriakidis, Matthew Purver, Ruth Kempson, Ronnie Cann, and
  Patrick G.~T. Healey. 2020.
\newblock \href {https://doi.org/10.1080/03740463.2020.1795549} {Completability
  vs (in)completeness}.
\newblock \emph{Acta Linguistica Hafniensia}, 52(2):260--284.

\bibitem[{Guhe(2007)}]{Guhe07}
Markus Guhe. 2007.
\newblock \emph{Incremental Conceptualization for Language Production}.
\newblock NJ: Lawrence Erlbaum Associates.

\bibitem[{Healey et~al.(2011)Healey, Eshghi, Howes, and Purver}]{Healey.etal11}
P.~G.~T. Healey, Arash Eshghi, Christine Howes, and Matthew Purver. 2011.
\newblock \href
  {http://www.eecs.qmul.ac.uk/~mpurver/papers/healey-et-al11std.pdf} {Making a
  contribution: Processing clarification requests in dialogue}.
\newblock In \emph{Proceedings of the 21st Annual Meeting of the Society for
  Text and Discourse}, Poitiers.

\bibitem[{Heldner et~al.(2013)Heldner, Hjalmarsson, and
  Edlund}]{Heldner.etal13}
Mattias Heldner, Anna Hjalmarsson, and Jens Edlund. 2013.
\newblock \href
  {http://www.diva-portal.org/smash/get/diva2:628698/FULLTEXT02.pdf}
  {Backchannel relevance spaces}.
\newblock In \emph{Nordic Prosody: Proceedings of XIth Conference, Tartu 2012},
  pages 137--146.

\bibitem[{Hough(2015)}]{Hough15}
Julian Hough. 2015.
\newblock \emph{Modelling Incremental Self-Repair Processing in Dialogue}.
\newblock Ph.D. thesis, Queen Mary University of London.

\bibitem[{Hough and Purver(2012)}]{Hough.Purver12}
Julian Hough and Matthew Purver. 2012.
\newblock \href
  {http://www.eecs.qmul.ac.uk/~mpurver/papers/hough-purver12semdial.pdf}
  {Processing self-repairs in an incremental type-theoretic dialogue system}.
\newblock In \emph{Proceedings of the 16th {SemDial} Workshop on the Semantics
  and Pragmatics of Dialogue ({SeineDial})}, pages 136--144, Paris, France.

\bibitem[{Howes and Eshghi(2021)}]{Howes.Eshghi21}
Christine Howes and Arash Eshghi. 2021.
\newblock \href {https://doi.org/10.1007/s10849-020-09328-1} {Feedback
  relevance spaces: Interactional constraints on processing contexts in dynamic
  syntax}.
\newblock \emph{Journal of Logic, Language and Information}, 30(2):331--362.

\bibitem[{Kahardipraja et~al.(2021{\natexlab{a}})Kahardipraja, Madureira, and
  Schlangen}]{Kahardipraja.21}
Patrick Kahardipraja, Brielen Madureira, and David Schlangen.
  2021{\natexlab{a}}.
\newblock \href {https://doi.org/10.18653/v1/2021.emnlp-main.90} {Towards
  incremental transformers: An empirical analysis of transformer models for
  incremental nlu}.
\newblock In \emph{Proceedings of the 2021 Conference on Empirical Methods in
  Natural Language Processing}, pages 1178--1189. Association for Computational
  Linguistics.

\bibitem[{Kahardipraja et~al.(2021{\natexlab{b}})Kahardipraja, Madureira, and
  Schlangen}]{Kahardipraja.etal21}
Patrick Kahardipraja, Brielen Madureira, and David Schlangen.
  2021{\natexlab{b}}.
\newblock \href {https://doi.org/10.18653/v1/2021.emnlp-main.90} {Towards
  incremental transformers: An empirical analysis of transformer models for
  incremental {NLU}}.
\newblock In \emph{Proceedings of the 2021 Conference on Empirical Methods in
  Natural Language Processing}, pages 1178--1189, Online and Punta Cana,
  Dominican Republic. Association for Computational Linguistics.

\bibitem[{Kahardipraja et~al.(2023)Kahardipraja, Madureira, and
  Schlangen}]{Kahardipraja.etal23}
Patrick Kahardipraja, Brielen Madureira, and David Schlangen. 2023.
\newblock \href {https://aclanthology.org/2023.findings-acl.257} {{TAPIR}:
  Learning adaptive revision for incremental natural language understanding
  with a two-pass model}.
\newblock In \emph{Findings of the Association for Computational Linguistics:
  ACL 2023}, pages 4173--4197, Toronto, Canada. Association for Computational
  Linguistics.

\bibitem[{Kalatzis et~al.(2016)Kalatzis, Eshghi, and Lemon}]{Kalatzis.etal16}
Dimitrios Kalatzis, Arash Eshghi, and Oliver Lemon. 2016.
\newblock Bootstrapping incremental dialogue systems: using linguistic
  knowledge to learn from minimal data.
\newblock In \emph{Proceedings of the NIPS 2016 workshop on {L}earning
  {M}ethods for {D}ialogue}, Barcelona.

\bibitem[{Katharopoulos et~al.(2020)Katharopoulos, Vyas, Pappas, and
  Fleuret}]{Katharopoulos.etal20}
Angelos Katharopoulos, Apoorv Vyas, Nikolaos Pappas, and Fran{\c{c}}ois
  Fleuret. 2020.
\newblock \href {https://proceedings.mlr.press/v119/katharopoulos20a.html}
  {Transformers are {RNN}s: Fast autoregressive transformers with linear
  attention}.
\newblock In \emph{Proceedings of the 37th International Conference on Machine
  Learning}, volume 119 of \emph{Proceedings of Machine Learning Research},
  pages 5156--5165. PMLR.

\bibitem[{Kempson et~al.(2015)Kempson, Cann, Eshghi, Gregoromichelaki, and
  Purver}]{Kempson.etal15}
Ruth Kempson, Ronnie Cann, Arash Eshghi, Eleni Gregoromichelaki, and Matthew
  Purver. 2015.
\newblock Ellipsis.
\newblock In Shalom Lappin and Chris Fox, editors, \emph{The Handbook of
  Contemporary Semantics}. Wiley-Blackwell.

\bibitem[{Kempson et~al.(2016)Kempson, Cann, Gregoromichelaki, and
  Chatzikiriakidis}]{Kempson.etal16}
Ruth Kempson, Ronnie Cann, Eleni Gregoromichelaki, and Stergios
  Chatzikiriakidis. 2016.
\newblock Language as mechanisms for interaction.
\newblock \emph{Theoretical Linguistics}, 42(3-4):203--275.

\bibitem[{Kempson et~al.(2001)Kempson, Meyer-Viol, and Gabbay}]{Kempson.etal01}
Ruth Kempson, Wilfried Meyer-Viol, and Dov Gabbay. 2001.
\newblock \emph{Dynamic Syntax: The Flow of Language Understanding}.
\newblock Wiley-Blackwell.

\bibitem[{Kennington et~al.(2014)Kennington, Kousidis, and
  Schlangen}]{Kennington.etal14}
Casey Kennington, Spyros Kousidis, and David Schlangen. 2014.
\newblock {InproTKs: A Toolkit for Incremental Situated Processing}.
\newblock In \emph{Proceedings of SIGdial 2014: Short Papers}, pages 84--88.

\bibitem[{Kwiatkowski et~al.(2012)Kwiatkowski, Goldwater, Zettlemoyer, and
  Steedman}]{Kwiatkowski.etal12}
Tom Kwiatkowski, Sharon Goldwater, Luke Zettlemoyer, and Mark Steedman. 2012.
\newblock A probabilistic model of syntactic and semantic acquisition from
  child-directed utterances and their meanings.
\newblock In \emph{Proceedings of the Conference of the European Chapter of the
  Association for Computational Linguistics (EACL)}.

\bibitem[{Larsson(2010)}]{Larsson10}
Staffan Larsson. 2010.
\newblock Accommodating innovative meaning in dialogue.
\newblock \emph{Proc. of Londial, SemDial Workshop}, pages 83--90.

\bibitem[{Larsson(2013)}]{Larsson13}
Staffan Larsson. 2013.
\newblock Formal semantics for perceptual classification.
\newblock \emph{Journal of logic and computation}.

\bibitem[{Levelt(1989)}]{Levelt89}
W.J.M. Levelt. 1989.
\newblock \emph{Speaking: From Intention to Articulation}.
\newblock {MIT Press}.

\bibitem[{MacWhinney(2000)}]{MacWhinney00}
Brian MacWhinney. 2000.
\newblock \emph{The {CHILDES} Project: Tools for Analyzing Talk}, third
  edition.
\newblock Lawrence Erlbaum Associates, Mahwah, New Jersey.

\bibitem[{Madureira and Schlangen(2020)}]{Madureira.Schlangen20}
Brielen Madureira and David Schlangen. 2020.
\newblock \href {https://doi.org/10.18653/v1/2020.emnlp-main.26} {Incremental
  processing in the age of non-incremental encoders: An empirical assessment of
  bidirectional models for incremental {NLU}}.
\newblock In \emph{Proceedings of the 2020 Conference on Empirical Methods in
  Natural Language Processing (EMNLP)}, pages 357--374, Online. Association for
  Computational Linguistics.

\bibitem[{Mao et~al.(2021)Mao, Shi, Wu, Levy, and Tenenbaum}]{Mao.etal21}
Jiayuan Mao, Freda~H. Shi, Jiajun Wu, Roger~P. Levy, and Joshua~B. Tenenbaum.
  2021.
\newblock \href {https://openreview.net/forum?id=VJQMp5xu24} {Grammar-based
  grounded lexicon learning}.
\newblock In \emph{Advances in Neural Information Processing Systems}.

\bibitem[{Nasreen et~al.(2021)Nasreen, Rohanian, Hough, and
  Purver}]{Nasreen.etal21}
Shamila Nasreen, Morteza Rohanian, Julian Hough, and Matthew Purver. 2021.
\newblock \href {https://doi.org/10.3389/fcomp.2021.640669} {Alzheimer’s
  dementia recognition from spontaneous speech using disfluency and
  interactional features}.
\newblock \emph{Frontiers in Computer Science}, 3.

\bibitem[{Nikolaus et~al.(2019)Nikolaus, Abdou, Lamm, Aralikatte, and
  Elliott}]{Nikolaus.etal19}
Mitja Nikolaus, Mostafa Abdou, Matthew Lamm, Rahul Aralikatte, and Desmond
  Elliott. 2019.
\newblock \href {https://doi.org/10.18653/v1/K19-1009} {Compositional
  generalization in image captioning}.
\newblock In \emph{Proceedings of the 23rd Conference on Computational Natural
  Language Learning (CoNLL)}, pages 87--98, Hong Kong, China. Association for
  Computational Linguistics.

\bibitem[{Pantazopoulos et~al.(2022)Pantazopoulos, Suglia, and
  Eshghi}]{Pantazopoulos.etal22}
George Pantazopoulos, Alessandro Suglia, and Arash Eshghi. 2022.
\newblock \href {https://doi.org/10.18653/v1/2022.acl-srw.11} {Combine to
  describe: Evaluating compositional generalization in image captioning}.
\newblock In \emph{Proceedings of the 60th Annual Meeting of the Association
  for Computational Linguistics: Student Research Workshop}, pages 115--131.
  Association for Computational Linguistics.

\bibitem[{Poesio and Rieser(2010)}]{Poesio.Rieser10}
Massimo Poesio and Hannes Rieser. 2010.
\newblock \href
  {http://elanguage.net/journals/index.php/dad/article/view/91/512}
  {Completions, coordination, and alignment in dialogue}.
\newblock \emph{Dialogue and Discourse}, 1:1--89.

\bibitem[{Purver et~al.(2011)Purver, Eshghi, and Hough}]{Purver.etal11}
Matthew Purver, Arash Eshghi, and Julian Hough. 2011.
\newblock Incremental semantic construction in a dialogue system.
\newblock In \emph{Proceedings of the 9th International Conference on
  Computational Semantics}, pages 365--369, Oxford, UK.

\bibitem[{Purver et~al.(2010)Purver, Gregoromichelaki, Meyer-Viol, and
  Cann}]{Purver.etal10}
Matthew Purver, Eleni Gregoromichelaki, Wilfried Meyer-Viol, and Ronnie Cann.
  2010.
\newblock Splitting the `{I}'s and crossing the `{Y}ou's: Context, speech acts
  and grammar.
\newblock In \emph{Aspects of Semantics and Pragmatics of Dialogue. SemDial
  2010, 14th Workshop on the Semantics and Pragmatics of Dialogue}, pages
  43--50, Pozna{\'n}. Polish Society for Cognitive Science.

\bibitem[{Purver et~al.(2009)Purver, Howes, Gregoromichelaki, and
  Healey}]{Purver.etal09}
Matthew Purver, Christine Howes, Eleni Gregoromichelaki, and Patrick G.~T.
  Healey. 2009.
\newblock \href
  {http://www.dcs.qmul.ac.uk/~mpurver/papers/purver-et-al09sigdial-corpus.pdf}
  {Split utterances in dialogue: {A} corpus study}.
\newblock In \emph{Proceedings of the 10th Annual {SIGDIAL} Meeting on
  Discourse and Dialogue ({SIGDIAL 2009} Conference)}, pages 262--271, London,
  UK. Association for Computational Linguistics.

\bibitem[{Purver and Kempson(2004)}]{Purver.Kempson04}
Matthew Purver and Ruth Kempson. 2004.
\newblock Incremental context-based generation for dialogue.
\newblock In \emph{Proceedings of the 3rd International Conference on Natural
  Language Generation ({INLG04})}, number 3123 in Lecture Notes in Artifical
  Intelligence, pages 151--160, Brockenhurst, UK. Springer.

\bibitem[{Purver et~al.(2021)Purver, Sadrzadeh, Kempson, Wijnholds, and
  Hough}]{Purver.etal21}
Matthew Purver, Mehrnoosh Sadrzadeh, Ruth Kempson, Gijs Wijnholds, and Julian
  Hough. 2021.
\newblock \href {https://doi.org/10.1007/s10849-021-09337-8} {Incremental
  composition in distributional semantics}.
\newblock \emph{Journal of Logic, Language and Information}, 30(2):379--406.

\bibitem[{Rashkin et~al.(2021)Rashkin, Reitter, Tomar, and
  Das}]{Rashkin.etal21}
Hannah Rashkin, David Reitter, Gaurav~Singh Tomar, and Dipanjan Das. 2021.
\newblock \href {https://doi.org/10.18653/v1/2021.acl-long.58} {Increasing
  faithfulness in knowledge-grounded dialogue with controllable features}.
\newblock In \emph{Proceedings of the 59th Annual Meeting of the Association
  for Computational Linguistics and the 11th International Joint Conference on
  Natural Language Processing (Volume 1: Long Papers)}, pages 704--718, Online.
  Association for Computational Linguistics.

\bibitem[{Rohanian and Hough(2021)}]{Rohanian.Hough21}
Morteza Rohanian and Julian Hough. 2021.
\newblock \href {https://doi.org/10.18653/v1/2021.acl-long.286} {Best of both
  worlds: Making high accuracy non-incremental transformer-based disfluency
  detection incremental}.
\newblock In \emph{Proceedings of the 59th Annual Meeting of the Association
  for Computational Linguistics and the 11th International Joint Conference on
  Natural Language Processing (Volume 1: Long Papers)}, pages 3693--3703,
  Online. Association for Computational Linguistics.

\bibitem[{Sato(2011)}]{Sato11}
Yo~Sato. 2011.
\newblock Local ambiguity, search strategies and parsing in {D}ynamic {S}yntax.
\newblock In E.~Gregoromichelaki, R.~Kempson, and C.~Howes, editors, \emph{The
  Dynamics of Lexical Interfaces}. CSLI Publications.

\bibitem[{Schegloff et~al.(1977)Schegloff, Jefferson, and
  Sacks}]{Schegloff.etal77}
E.A. Schegloff, Gail Jefferson, and Harvey Sacks. 1977.
\newblock {The preference for self-correction in the organization of repair in
  conversation}.
\newblock \emph{Language}, 53(2):361--382.

\bibitem[{Schlangen and Skantze(2009)}]{Schlangen.Skantze09}
David Schlangen and Gabriel Skantze. 2009.
\newblock \href {http://www.aclweb.org/anthology/E09-1081} {A general, abstract
  model of incremental dialogue processing}.
\newblock In \emph{Proceedings of the 12th Conference of the European Chapter
  of the ACL (EACL 2009)}, pages 710--718, Athens, Greece. Association for
  Computational Linguistics.

\bibitem[{Skantze and Hjalmarsson(2010)}]{Skantze.Hjalmarsson10}
Gabriel Skantze and Anna Hjalmarsson. 2010.
\newblock \href
  {http://www.sigdial.org/workshops/workshop11/proc/pdf/SIGDIAL01.pdf} {Towards
  incremental speech generation in dialogue systems}.
\newblock In \emph{Proceedings of the SIGDIAL 2010 Conference}, pages 1--8,
  Tokyo, Japan. Association for Computational Linguistics.

\bibitem[{Skantze and Schlangen(2009)}]{Skantze.Schlangen09}
Gabriel Skantze and David Schlangen. 2009.
\newblock \href {http://www.aclweb.org/anthology/E09-1085} {Incremental
  dialogue processing in a micro-domain}.
\newblock In \emph{Proceedings of the 12th Conference of the European Chapter
  of the ACL (EACL 2009)}, pages 745--753, Athens, Greece. Association for
  Computational Linguistics.

\bibitem[{Vaswani et~al.(2017)Vaswani, Shazeer, Parmar, Uszkoreit, Jones,
  Gomez, Łukasz Kaiser, and Polosukhin}]{Vaswani.etal17}
Ashish Vaswani, Noam Shazeer, Niki Parmar, Jakob Uszkoreit, Llion Jones,
  Aidan~N. Gomez, Łukasz Kaiser, and Illia Polosukhin. 2017.
\newblock Attention is all you need.
\newblock In \emph{Advances in Neural Information Processing Systems}, volume
  2017-December.

\bibitem[{Yu et~al.(2016)Yu, Eshghi, and Lemon}]{Yu.etal16}
Yanchao Yu, Arash Eshghi, and Oliver Lemon. 2016.
\newblock Training an adaptive dialogue policy for interactive learning of
  visually grounded word meanings.
\newblock In \emph{Proceedings of SIGDIAL 2016, 17th Annual Meeting of the
  Special Interest Group on Discourse and Dialogue}, pages 339--349, Los
  Angeles.

\bibitem[{Yu et~al.(2017)Yu, Eshghi, and Lemon}]{Yu.etal17b}
Yanchao Yu, Arash Eshghi, and Oliver Lemon. 2017.
\newblock \href {http://www.aclweb.org/anthology/W17-2802} {Learning how to
  learn: an adaptive dialogue agent for incrementally learning visually
  grounded word meanings}.
\newblock In \emph{Proceedings of the {F}irst {W}orkshop on {L}anguage
  {G}rounding for {R}obotics}, pages 1--10. ACL.

\end{thebibliography}
\bibliographystyle{acl_natbib}




\end{document}